\documentclass{article}

\usepackage[T1]{fontenc}
\usepackage{graphicx}
\usepackage{subcaption}
\usepackage{booktabs} 
\usepackage{amsmath}
\usepackage{amssymb}
\usepackage{mathtools}
\usepackage{amsthm}
\usepackage{natbib}

\usepackage{hyperref}
\usepackage{algorithmic}

\theoremstyle{plain}

\theoremstyle{definition}

\theoremstyle{remark}

\DeclareMathOperator*{\argmin}{arg\,min}

\title{Domain Adaptation Under Wireless Network Constraints: When Does It Become Green?}

\author{
Illyyne Saffar\textsuperscript{1,*},
Aurelie Boisbunon\textsuperscript{1},
Shruti Bothe\textsuperscript{2}
\\
\\
\bigskip
\bf{Ericsson Research}\\
\textsuperscript{1}Paris, France \quad
\textsuperscript{2}Santa Clara, USA\\
\bigskip
\normalsize
* Corresponding author: illyyne.saffar@ericsson.com \\
}

\date{}

\begin{document}

\maketitle

\begin{abstract}
  The deployment of data-driven models in 6G wireless networks is increasingly challenged by frequent distribution shifts that degrade performance over time. Unsupervised Domain Adaptation (UDA) offers an  alternative approach by adapting the trained model to a shifted domain without requiring labels. However, UDA pipelines are often more complex than single-task training due to additional modules and optimization procedures, raising a practical question: do the benefits of adaptation come at a higher energy cost, and how does this trade-off compare to retraining when labeling effort is also considered?
  In this work, we investigate the energy consumption of UDA and compare it to single task. 
  We further propose a way to determine the minimum number of target domains for which UDA becomes more energy-efficient than retraining, taking into account the labeling cost. 
  Our results aim to clarify when UDA should be preferred over classical train-from-scratch approaches from an energy and labeling-aware perspective.
\end{abstract}

\section{Introduction}

The sixth generation (6G) of wireless communication is expected to have a tenfold increase in traffic load, enabling higher connectivity and better services~\citep{ericsson_mobility_report}. However, this growth is {accompanied} by extreme dynamism and heterogeneity across deployment scenarios: cell densities, traffic patterns, channel models, and user mobility vary significantly {in different} environments. As a result, data‑driven models trained under {a single} set of conditions {such as} a specific urban micro‑cell with a particular fading profile, consistently fails to generalize to other deployments, a phenomenon known as domain shift. Large generative models like foundation models can in principle bridge this generalization gap, but their high inference latency (often hundreds of milliseconds to hundreds of seconds) conflicts with the strict real‑time requirement constraints of many wireless tasks, such as channel estimation, interference classification, and handover decisions, which demand sub‑second to microsecond responses~\citep{maatouk2024large,lin2022device}. Consequently, such large models are impractical for edge and base‑station deployment.

Unsupervised Domain Adaptation (UDA) offers a promising alternative by enabling relatively small models, trained on abundant labeled data from a source domain (e.g., simulations or laboratory environments), to adapt to an unlabeled target domain (e.g., a live rural or dense‑urban deployment) without requiring expensive ground‑truth annotations in the field~\cite{fawaz2025deep}. For wireless applications where signal‑to‑noise ratio, multipath fading, and user mobility vary substantially, UDA can dramatically reduce the need for site‑specific data collection and labeling, thereby accelerating the scalable deployment of AI‑driven systems. Yet, wireless networks impose additional practical constraints beyond domain shift: models must be lightweight for continuous inference, amenable to rapid retraining on short timescales, 
and energy‑efficient to support the sustainability goals that 6G explicitly commits to~\cite{ericsson_etr_when_ai_has_no_time_to_think, nextgalliance2025sustainable}. These requirements challenge conventional assumptions about domain adaptation, which has largely been studied in computer vision and natural language processing where latency and energy budgets are far less restrictive.

In this work, we evaluate the adaptability of time‑series classification tasks under realistic wireless conditions, with a particular focus on measuring algorithmic efficiency from a green AI perspective. Intuitively, avoiding full retraining from scratch and bypassing costly labeling sounds more efficient, but whether UDA actually delivers a favorable trade‑off among energy consumption, labeling cost, adaptation speed, and generalization performance remains an open question, both theoretically and empirically. Our analysis aims to check where UDA outperforms both training from scratch (expensive in energy and labels) and fine‑tuning large pre‑trained models (too slow and power‑hungry), providing practical guidelines for deploying sustainable, real‑time UDA in next‑generation wireless systems.

\section{Context and notations}

\subsection{Wireless Network Constraints} 

Wireless networks present a unique set of constraints that challenge conventional AI pipelines. First, wireless data is inherently \textbf{temporal} in nature, with 
correlations and non-stationarities that standard i.i.d. assumptions violate~\citep{o2017introduction,sun2018learning}. 
Second, \textbf{label scarcity} is a recurrent issue, as the ground-truth annotation for many use cases like channel states, interference sources, or traffic classes in live deployments is expensive and often impractical, on top of which it raises privacy aspects~\citep{soares2023semi,saffar2019semi}. 
Third, \textbf{domain shifts} frequently arise from changes in the {spatio-temporal} environment (user mobility, weather, network load, geographical deployment, countries, or hardware configurations), occurring far more often than in typical computer vision or NLP domains.  In particular, \textbf{covariate shift} is frequently encountered in wireless networks due to their dynamicity \citep{raza2014adaptive,raghuram2021few,talak2018optimizing}.
Fourth, any practical solution must be \textbf{lightweight for inference and amenable to rapid or offline retraining} on short time scales (micro- or milliseconds) to comply with operational constraints, ruling out large foundation models or slow online learning.
Finally, there is a need for \textbf{sustainable AI}~\citep{ericsson_etr_when_ai_has_no_time_to_think,bothe2025through,nextgalliance2025sustainable,lin2022device, lai2026comprehensive}. Indeed, ICT sector used about $4\%$
of the global electricity in the use stage and represented about $1.4\%$ of the global GHG emissions in
2020~\citep{buzzi2016survey,malmodin2024ict}. It holds the potential to enable a $15\%$ reduction in cross-industry emissions by 2030 through connectivity-driven solutions such as smart and autonomous networks, smart building management and connected electric vehicle charging infrastructure~\citep{ericsson_sustainability_2023}. If we integrate AI to support these green networks, then AI itself must be sustainable in return. Models must therefore have low energy footprints during both training and inference. 

Domain shift and label scarcity correspond to the main assumptions of UDA techniques, which make such approaches particularly indicated for wireless applications. However, the application of UDA to time series presents unique challenges due to the inherent characteristics of such a data type.
Unlike images, time series data involves temporal ordering, trends, seasonality, and varying frequencies, all of which influence analysis and performance. These factors contribute to the increased complexity of UDA for time series. 
Additionally, To the best of our knowledge there are no previous work comparing the energy consumption of UDA with retraining from scratch standard classification approaches on new datasets. Within this realistic setting, we investigate the following central question: \textit{When does Unsupervised Domain Adaptation (UDA) become "green" and offer a favorable compromise among energy efficiency, labeling cost, adaptation speed, and generalization performance?} 
Our contributions identify the regimes in which UDA surpasses both training from scratch which incurs high energy and labeling costs and fine-tuning large pre-trained models, which can be slow and computationally intensive. Based on these findings, we offer practical guidelines for applying UDA in next-generation wireless systems.

\subsection{UDA for Time series classification}
Let $\mathbf{x}=(x^1, \dots,x^d)\in\mathbb{R}^{t\times d}$, be a time series where $d$ is the number of features (or channels) and $t$ its length. 
In classification, the goal is to predict $y \in \{0, \ldots, c-1\}$
from a labeled set $\{(\mathbf{x}_i, y_i)\}_{i=1}^{n}$ drawn i.i.d.\ from distribution $\mathcal{D}$.
In the telecom context, a representative example is traffic classification, where
inputs are features such as packet inter-arrival times and outputs are traffic
types, e.g., voice, data, SMS.

UDA aims to transfer knowledge from a labeled source domain to an unlabeled target domain by reducing the discrepancy (or also called shift) between their distributions, without access to target labels during training. This shift may occur at the level of the input distribution (covariate shift), the label distribution (label shift), or the relationship
between inputs and labels (concept drift). This paper focuses on covariate shift, following the mathematical setting of~\citep{ben2010}. Consider two distributions over $\mathcal{X} \times \mathcal{Y}$: a source domain $\mathcal{D}_S$ and a target domain $\mathcal{D}_T$. A UDA algorithm is provided with 
$n_S$ i.i.d. a labeled source dataset 
$S = \{(\mathbf{x}_i^S, y_i^S)\}_{i=1}^{n_S}$ drawn i.i.d.\ from $\mathcal{D}_S$, and an unlabeled target 
dataset $T = \{\mathbf{x}_i^T\}_{i=1}^{n_T}$ drawn i.i.d.\ from $\mathcal{D}_T^X$, the marginal distribution of $\mathcal{D}_T$ over $\mathcal{X}$. The goal is to learn a classifier $\eta: \mathcal{X} \rightarrow \mathcal{Y}$
with low target risk: 
$$
R_{\mathcal{D}_T}(\eta) =
\mathbb{P}_{(\mathbf{x}, y) \sim \mathcal{D}_T}\left[\eta(\mathbf{x}) \neq y\right],
$$
without access to target labels. Under the covariate shift assumption,
$p_S(\mathbf{x}) \neq p_T(\mathbf{x})$ while
$p_S(y \mid \mathbf{x}) = p_T(y \mid \mathbf{x})$.

\section{Energy Consumption UDA vs TSC}

\subsection{AI \& Carbon footprint estimation}
Quantifying the energy footprint of AI models is critical not only at training and inference time, but across the full AI lifecycle: data collection and preprocessing, model training, validation, inference, monitoring for
distribution shift, and retraining upon model staleness. Energy consumption is typically measured in kWh via hardware power sampling (e.g., RAPL for CPUs,
NVML for GPUs) or estimated from FLOPs and hardware TDP~\citep{patterson2021carbon}.
To relate energy to environmental impact, several open-source libraries support end-to-end carbon accounting within Python-based workflows, including CarbonTracker~\citep{carbontracker}, Eco2AI~\citep{eco2ai}, and CodeCarbon~\citep{courty2024codecarbon}. These tools estimate the carbon footprint $\text{CF}$ (in gCO$_2$eq) as: $\text{CF} = \text{CI} \times \text{PUE} \times E,$ where $\text{CI}$ (gCO$_2$eq/kWh) is the carbon intensity of the local electricity grid, $\text{PUE} \geq 1$ is the Power Usage Effectiveness of the compute infrastructure, and $E$ (kWh) is the total energy consumed by
the stage under consideration. This unified approach is very beneficial since it allows quantifying the different stages of the of the AI life cycle management (LCM-AI) on the same scale.

\begin{figure}[t]
    \centering
    \includegraphics[width=0.9\linewidth]{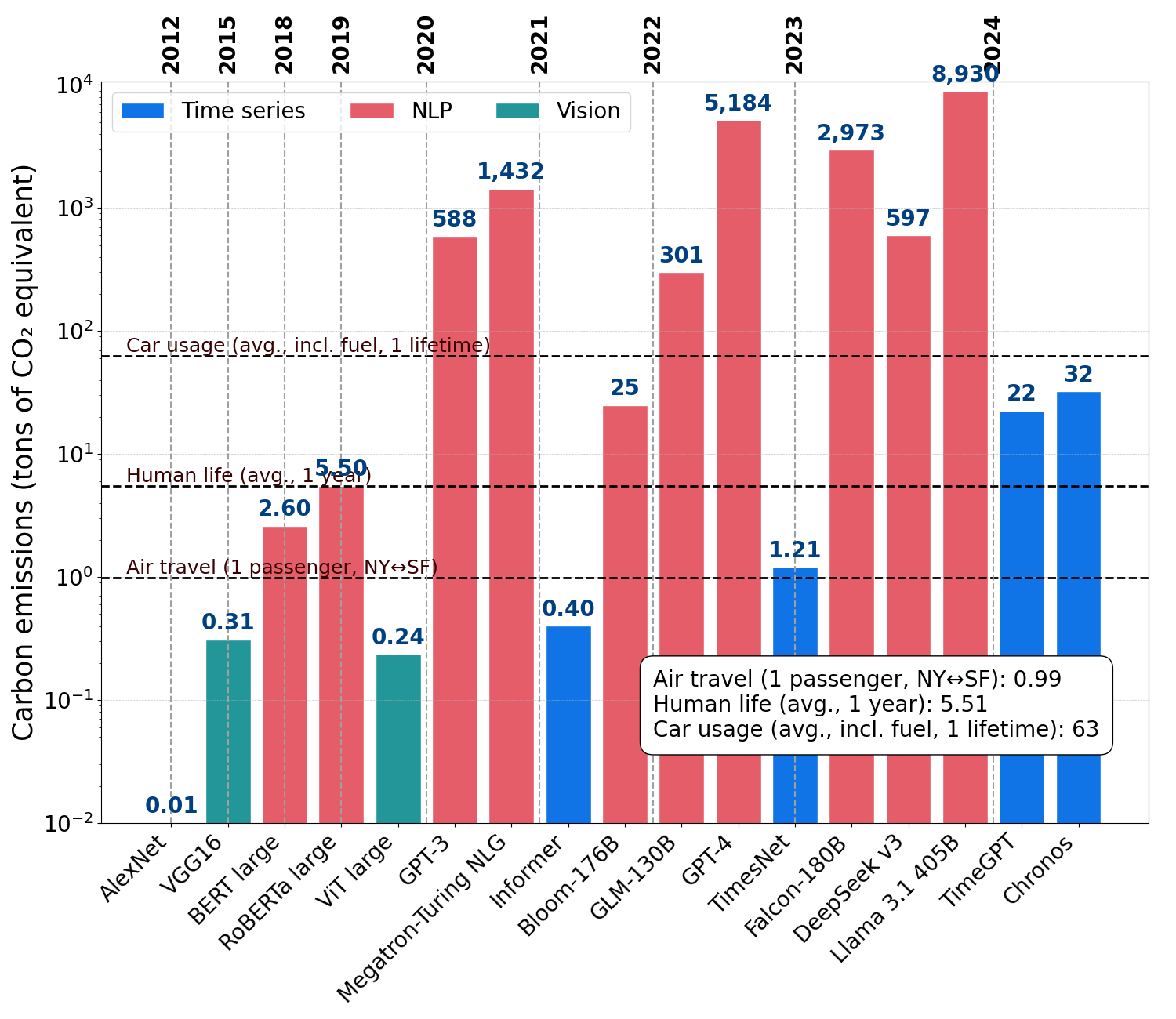}
    \caption{Estimated carbon emissions from training select AI models and real-life activities, 2012–24.}
    \label{fig:carbonEmissionModels}
\end{figure}

The scale of these costs varies dramatically across model families (see Fig.~\ref{fig:carbonEmissionModels}). 
Training a single BERT-base model emits approximately $2.6$ tons od CO$_2$eq, while TimeGPT can reach $22$ tons of CO$_2$eq~\citep{strubell2019energy,lacoste2019quantifying}. 
For some families of models, such as large language models (LLM), the total life cycle cost is dominated by the inference phase, using much higher resources ($65\%$) as compared to training ($35\%$) due to the millions of daily queries~\citep{desislavov2023trends,wu2022sustainable}.
In contrast, lightweight models such as logistic regression or small CNNs consume several orders of magnitude less energy, making model selection a critical lever for Green AI~\citep{mao2021ai}. These observations motivate our focus on relatively small, efficient UDA models suited to the resource-constrained wireless deployments.

\subsection{UDA versus Independent Classification: Cost Analysis and Break-Even Estimation}

To quantify the efficiency gains of UDA when adapting a single source model to
multiple target domains, we compare its computational and environmental cost
against an independent classification baseline, in which a separate supervised
model is trained from scratch for each target domain.
\paragraph{Pipeline Stages.}
Let $\mathcal{C} = \{c_1, c_2, c_3, c_4, c_5\}$ denote the ordered set of
pipeline stages, defined as follows:
\begin{align}
    c_1: &\quad \texttt{preprocess} \quad
        f_{\mathrm{pre}}: \mathcal{X}_{\mathrm{raw}} \rightarrow \mathcal{X},
        \quad \mathbf{x} \mapsto \tilde{\mathbf{x}} \\
    c_2: &\quad \texttt{tune} \quad
        f_{\mathrm{tune}}: \Lambda \rightarrow \Lambda^*, \nonumber \\
        & \hspace{4em}
        \lambda^* = \operatorname*{arg\,min}_{\lambda \in \Lambda}\,
        \mathcal{L}\!\left(f_{\mathrm{pred}}\!\circ
        f_{\mathrm{train}}(\lambda)\right)\\
    c_3: &\quad \texttt{train} \quad
        f_{\mathrm{train}}: \mathcal{X} \times \Lambda^* \rightarrow \mathcal{H},
        \quad (\tilde{\mathbf{x}}, \lambda^*) \mapsto h^* \\
    c_4: &\quad \texttt{predict} \quad
        f_{\mathrm{pred}}: \mathcal{X} \rightarrow \mathcal{Y},
        \quad \mathbf{x} \mapsto \hat{y} = h^*(\mathbf{x}) \\
    c_5: &\quad \texttt{score} \quad
        f_{\mathrm{score}}: \mathcal{Y} \times \mathcal{Y} \rightarrow \mathbb{R},
        \quad (\hat{y}, y) \mapsto \ell
\end{align}
where $\mathcal{X}_{\mathrm{raw}}$ is the raw input space, $\mathcal{X}$ the
preprocessed feature space, $\Lambda$ the hyperparameter search space,
$\Lambda^*$ the optimal hyperparameter configuration, $\mathcal{H}$ the
hypothesis class, and $\mathcal{Y}$ the label space.
Stage $c_2$ is implemented as a Ray Tune orchestration loop that
repeatedly calls $f_{\mathrm{train}}$ and $f_{\mathrm{pred}}$ until the
trial budget $B \in \mathbb{N}$ is exhausted, returning
$\lambda^* = \argmin_{\lambda} \mathcal{L}$. Stages $c_3$--$c_5$ are
then executed once with $\lambda^*$.
Each stage $c \in \mathcal{C}$ incurs a measurable energy cost:
$$
E^{(c)}(r) \in \mathbb{R}_{+},
$$
where $r$ denotes a run and energy is reported in kWh (carbon footprint
in gCO$_2$eq).

\paragraph{Setup and Notation.}
Let $S$ denote the source domain, $\{T_1, \ldots, T_N\}$ a set of $N$ target
domains, and $c \in \mathcal{C}$ a pipeline stage. The total cost of a
strategy $\pi$ over stage $c$ and $N$ target domains is denoted
$C_{\pi,\mathrm{total}}^{(c)}(N)$. Three strategies are compared:
\begin{itemize}
    \item \textbf{Independent Classification (IC):} for each target domain
    $T_i$, all stages $c \in \mathcal{C}$ are executed independently,
    including labeling and full retraining from scratch.
    \item \textbf{UDA -- Sequential Adaptation (UDA-S):} stages $c_1$--$c_3$
    are executed once on $S$; stages $c_1$ and $c_3$ (adaptation) are then
    repeated independently for each $T_i$, $i = 1, \ldots, N$.
    \item \textbf{UDA -- Joint Multi-Target Adaptation (UDA-J):} stages
    $c_1$--$c_3$ are executed once on $S$; a single adaptation run at stage
    $c_3$ is performed jointly over $\{T_1, \ldots, T_N\}$ simultaneously.
\end{itemize}

\paragraph{Independent Classification Cost Model.}
In the independent setting, stage $c_1$ incurs both a labeling cost and a
computational cost, while all other stages incur only computational cost.
The total per-domain energy cost at stage $c$ is:
$$
E_{\mathrm{ind}}^{(c)} =
\begin{cases}
    E_{\mathrm{label}}^{(c_1)} + E_{\mathrm{comp}}^{(c_1)} & \text{if } c = c_1, \\
    E_{\mathrm{comp}}^{(c)} & \text{otherwise,}
\end{cases}
$$
where the average computational cost per target domain at stage $c$ is estimated as:
$$
E_{\mathrm{comp}}^{(c)} = \frac{1}{N}
\sum_{i=1}^{N} \sum_{r \in \mathcal{R}_{T_i,c}} E^{(c)}(r),
$$
with $\mathcal{R}_{T_i,c}$  the set of runs executed for target domain
$T_i$ at stage $c$, and $E^{(c)}(r)$  the measured energy for run $r$.

The total cost over $N$ target domains thus scales linearly:
$$
E_{\mathrm{IC,total}}^{(c)}(N) = E_{\mathrm{source}}^{(c)} + N \cdot E_{\mathrm{ind}}^{(c)} ,
$$

\paragraph{UDA Cost Model: Sequential Adaptation (UDA-S).}
The source model is trained once and adapted independently to each $T_i$.
No labeling cost is incurred at any stage. The total cost is:
$$
E_{\mathrm{UDA\text{-}S,total}}^{(c)}(N) = E_{\mathrm{source}}^{(c)}
+ N \cdot E_{\mathrm{adapt}}^{(c)},
$$
where:
$$
E_{\mathrm{source}}^{(c)} = \sum_{r \in \mathcal{R}_{s,c}} E^{(c)}(r)$$
$$
E_{\mathrm{adapt}}^{(c)} = \frac{1}{|\mathcal{R}_{\mathrm{UDA},c}|}
\sum_{r \in \mathcal{R}_{\mathrm{UDA},c}} E^{(c)}(r).
$$

\paragraph{UDA Cost Model: Joint Adaptation (UDA-J).}
The source model is adapted once to all $N$ target domains simultaneously.
The total cost reduces to:
$$
E_{\mathrm{UDA\text{-}J,total}}^{(c)}(N) = E_{\mathrm{source}}^{(c)}
+ E_{\mathrm{adapt,joint}}^{(c)}(N),
$$
where $E_{\mathrm{adapt,joint}}^{(c)}(N)$ is the energy cost of a single
joint adaptation run over all $N$ targets. Since all domains are processed
in a single pass:
$$
E_{\mathrm{adapt,joint}}^{(c)}(N) \leq N \cdot E_{\mathrm{adapt}}^{(c)},
$$
with equality only when the joint run degenerates to $N$ sequential passes.

\paragraph{Break-Even Analysis.}
The break-even number of targets $N^*$ is the minimum $N$ beyond which a
UDA strategy becomes more energy-efficient than IC.

\noindent\textbf{UDA-S break-even:}
$$
N^*_{\mathrm{S}} = \frac{E_{\mathrm{source}}^{(c)}}
{E_{\mathrm{ind}}^{(c)} - E_{\mathrm{adapt}}^{(c)}},
\qquad \text{provided } E_{\mathrm{adapt}}^{(c)} < E_{\mathrm{ind}}^{(c)}.
$$

\noindent\textbf{UDA-J break-even:}
$$
N^*_{\mathrm{J}} = \frac{E_{\mathrm{source}}^{(c)}
+ E_{\mathrm{adapt,joint}}^{(c)}}{E_{\mathrm{ind}}^{(c)}}.
$$
Since $E_{\mathrm{adapt,joint}}^{(c)}$ is fixed regardless of $N$, it
follows that $N^*_{\mathrm{J}} \leq N^*_{\mathrm{S}}$, confirming that
joint adaptation reaches the efficiency threshold earlier. The three
strategies are summarized in Table~\ref{tab:cost_summary}.

\begin{table}[h]
\centering
\caption{Summary of total energy cost models and break-even thresholds. $(^{*})$ For IC, the $(N+1)$ factor accounts for the source domain.}
\label{tab:cost_summary}
\scriptsize
\setlength{\tabcolsep}{0.5pt}
\begin{tabular}{lccc}
\toprule
\textbf{Strategy} & \textbf{Total Cost $E^{(c)}$} & \textbf{Scales with $N$}
& \textbf{Break-Even $N^*$} \\
\midrule
IC$^{*}$
    & $(N+1) \cdot (E_{\mathrm{label}}^{(c_1)} + E_{\mathrm{comp}}^{(c)})$
    & Linear & --- \\
UDA-S
    & $E_{\mathrm{source}}^{(c)} + N \cdot E_{\mathrm{adapt}}^{(c)}$
    & Linear
    & $\frac{E_{\mathrm{source}}^{(c)}}{E_{\mathrm{ind}}^{(c)} - E_{\mathrm{adapt}}^{(c)}}$ \\
UDA-J
    & $E_{\mathrm{source}}^{(c)} + E_{\mathrm{adapt,joint}}^{(c)}$
    & Sublinear
    & $\frac{E_{\mathrm{source}}^{(c)} + E_{\mathrm{adapt,joint}}^{(c)}}{E_{\mathrm{ind}}^{(c)}}$ \\
\bottomrule
\end{tabular}
\end{table}

\section{Experiments}

\subsection{Datasets}
To the best of our knowledge, no open-source dataset exists for UDA-based time series classification in the wireless or telecommunications domain. We therefore follow~\citet{fawaz2025deep} and select benchmarks from adjacent domains exhibiting the distribution shifts and label scarcity characteristic of real-world wireless deployments, spanning three application areas. \textbf{Human activity recognition:} HAR~\citep{anguita2013public}, HHAR~\citep{stisen2015smart}, and WISDM~\citep{kwapisz2011activity} provide multivariate wearable sensor signals (accelerometers, gyroscopes) with the task of inferring activity type. \textbf{Sports and daily-life motion:} five body-worn sensors record 19 activities from 8 subjects~\citep{altun2010comparative}, with standard windowing, denoising, and normalization applied. \textbf{Mechanical fault diagnosis:} the CWRU bearing dataset~\citep{bearingdataset} contains vibration signals from motor bearings under normal and faulty conditions, preprocessed via detrending, band-pass filtering, and segmentation.

\subsection{Models and experimental setup}

The experimental setup used here is closely related to 
the UDA benchmark performed in \cite{fawaz2025deep}, where the following UDA approaches representative of current practice are compared.

\textbf{Adversarial representation alignment}: methods that train a feature extractor to produce domain-invariant representations by adversarial confusing a domain discriminator. 

\textbf{Discrepancy minimization}: methods that minimize explicit measures of distribution mismatch (for example, kernel-based distances such as maximum mean discrepancy) or that employ specialized convolutional/encoder backbones designed for time series.

\textbf{Strong representation learners}: encoder architectures (e.g., Inception-like encoders) used in combination with adaptation objectives; these often attain high accuracy but can be computationally more demanding.

\textbf{Task-independent classification baseline}: Since the
Inception architecture consistently emerged as the strongest backbone across
UDA experiments~\cite{fawaz2025deep}, InceptionTime was selected as the task-independent baseline.

For each run, energy consumption is measured using CodeCarbon~\citep{courty2024codecarbon}, and all cost estimates are reported
as empirical means over MLflow-tracked runs. All experiments are implemented in PyTorch and executed using the Ray
framework for distributed hyperparameter tuning and experiment orchestration. To ensure a fair comparison across all methods, we adopt a fixed time budget for both the tuning stage ($c_2$) and the training stage ($c_3$), held constant across all models and datasets. This budget-based protocol prevents any single method from benefiting from disproportionate tuning or training resources, and ensures that the reported energy costs are directly comparable across strategies.

\begin{figure}[t]
    \centering
    \begin{subfigure}[b]{0.49\linewidth}
        \centering
        \includegraphics[width=\linewidth]{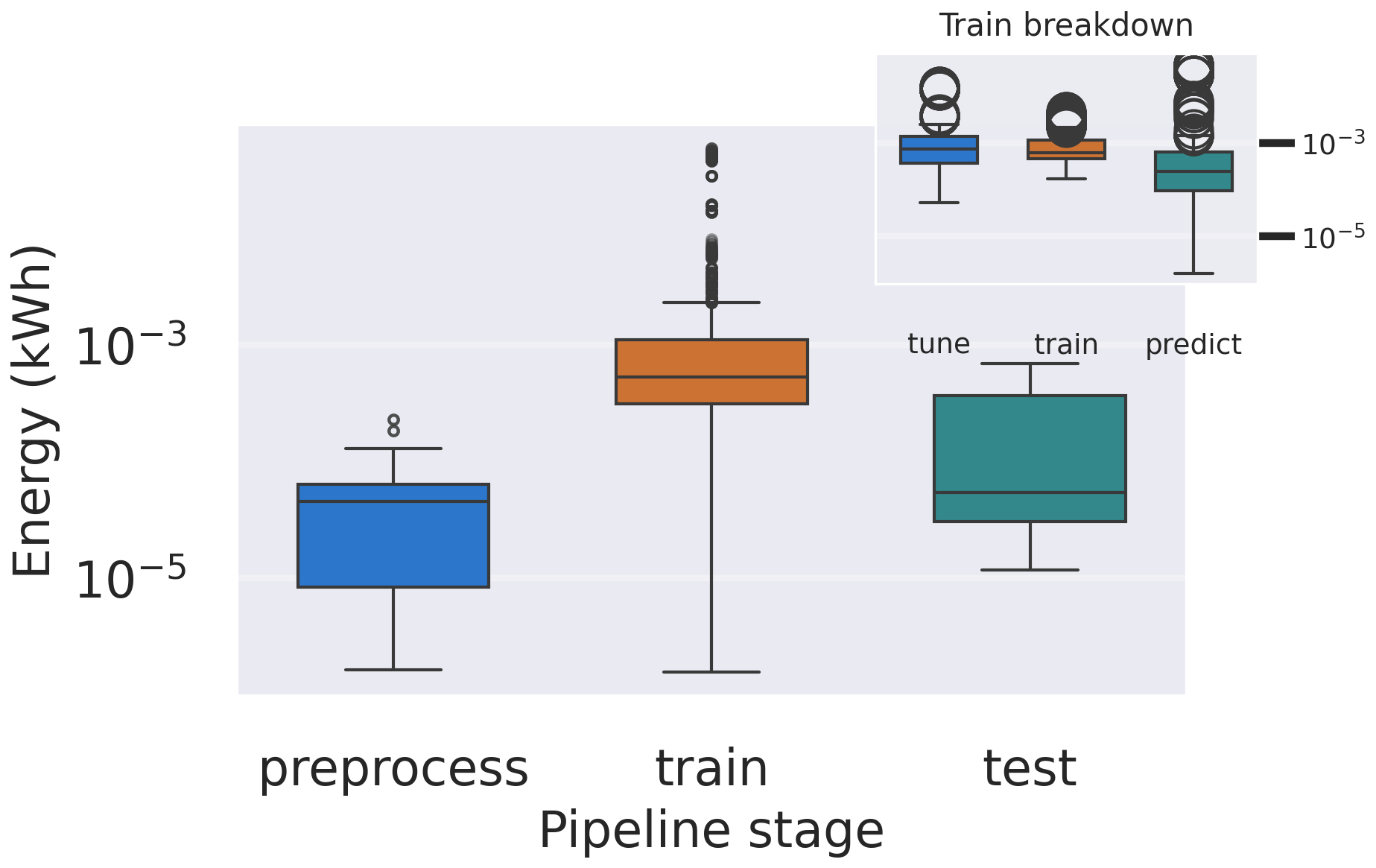}
        \caption{UDA}
        \label{fig:energy-uda}
    \end{subfigure}
    \hfill
    \begin{subfigure}[b]{0.49\linewidth}
        \centering
        \includegraphics[width=\linewidth]{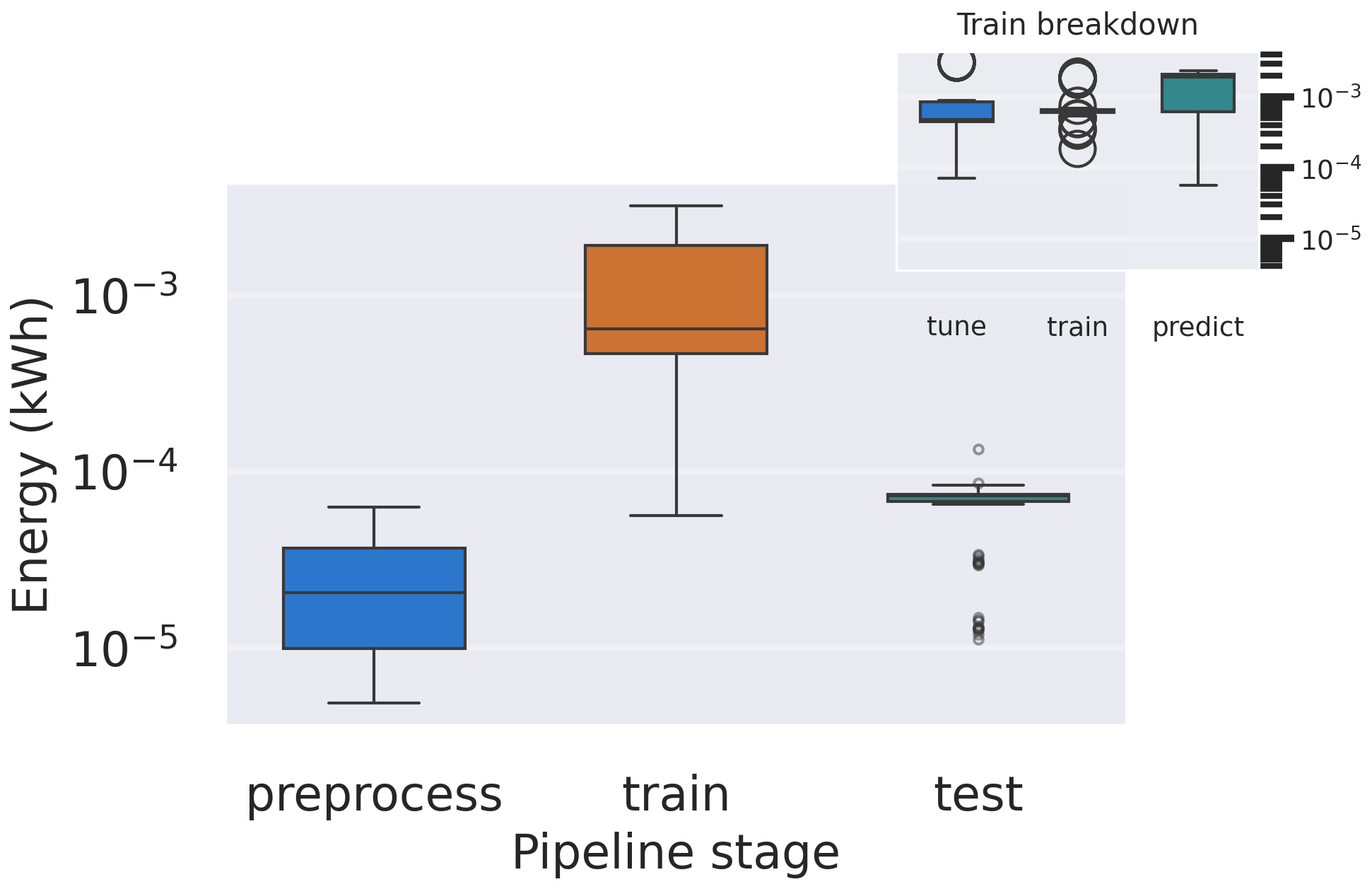}
        \caption{Independent classification}
        \label{fig:energy-clf}
    \end{subfigure}
    \caption{Per-stage energy consumption (kWh) on a logarithmic scale for the UDA pipeline (\emph{left}) and the independent classification pipeline (\emph{right}). Each box reports the distribution over all experimental runs. The inset details the breakdown of the training stage into its three sub-components: hyperparameter search (\texttt{tune}), final model fitting (\texttt{train}), and inference (\texttt{predict}).}
    \label{fig:energy}
\end{figure}
\subsection{Results}

\paragraph{Energy consumption per stage.} Figure~\ref{fig:energy} breaks down the estimated energy consumption by pipeline stage
for (a) the UDA and (b) the independent classification settings.
The training stage dominates by two to three orders of magnitude over preprocessing and inference.
This is explained by the pipeline structure: the tuning phase
(\texttt{tune}) repeatedly calls \texttt{train} and \texttt{predict}
until the hyperparameter budget is exhausted, after which a final
\texttt{train} run is performed with the best configuration found,
followed by a single \texttt{score} pass. Consequently, \texttt{tune} concentrates the bulk of the total energy and drives most of the run-to-run variance, while the final \texttt{train} and \texttt{score} steps contribute a comparatively negligible share. The UDA pipeline exhibits a higher upper tail than the classification setting, reflecting the sensitivity of distribution-matching objectives. Whether UDA or independent classification is the greener choice overall cannot be determined from this view alone.

\begin{figure}[ht!]
    \centering
    \includegraphics[width=0.6\linewidth]{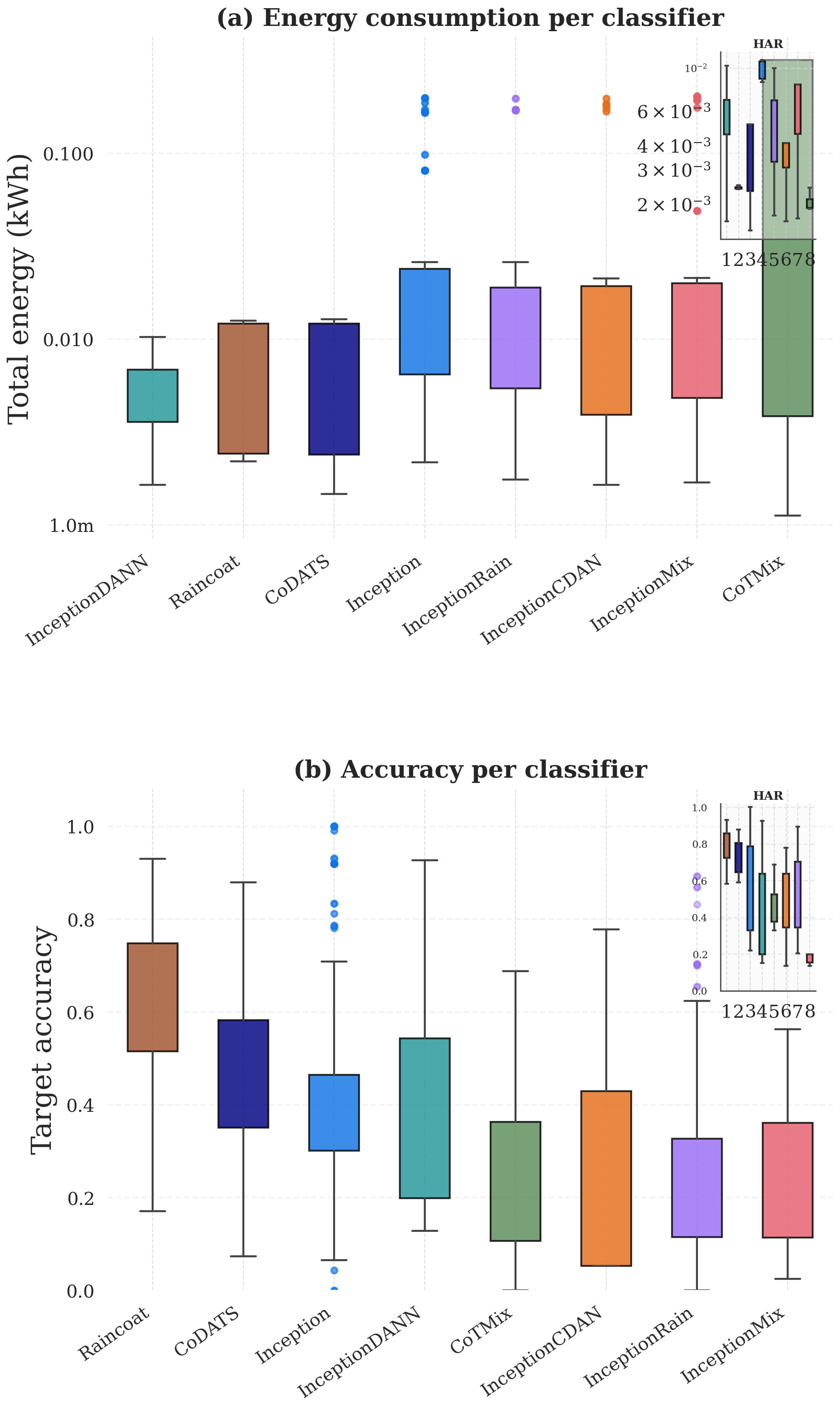}
    \caption{Comparison of energy consumption (KWh, top) and performance (bottom) across algorithms.}
    \label{fig:energy_alg_comp}
\end{figure}

\paragraph{Energy consumption per algorithm.} Figure~\ref{fig:energy_alg_comp} jointly characterizes the energy-accuracy profile of UDA classifiers. Figure~(a) shows that InceptionDANN, Raincoat, and CoDATS remain the cheapest group (median below $10^{-2}$\,kWh), while Inception-based UDA variants consume roughly $15$--$20$\,m\,kWh at median. CoTMix stands apart with an inter-quartile range spanning nearly two orders of magnitude, reflecting highly variable convergence across transfer pairs. The upper right plots show a zoom for the HAR dataset, which confirms that these relative rankings are stable, with all models.

\begin{figure*}[ht!]
  \centering
  \includegraphics[width=\linewidth]{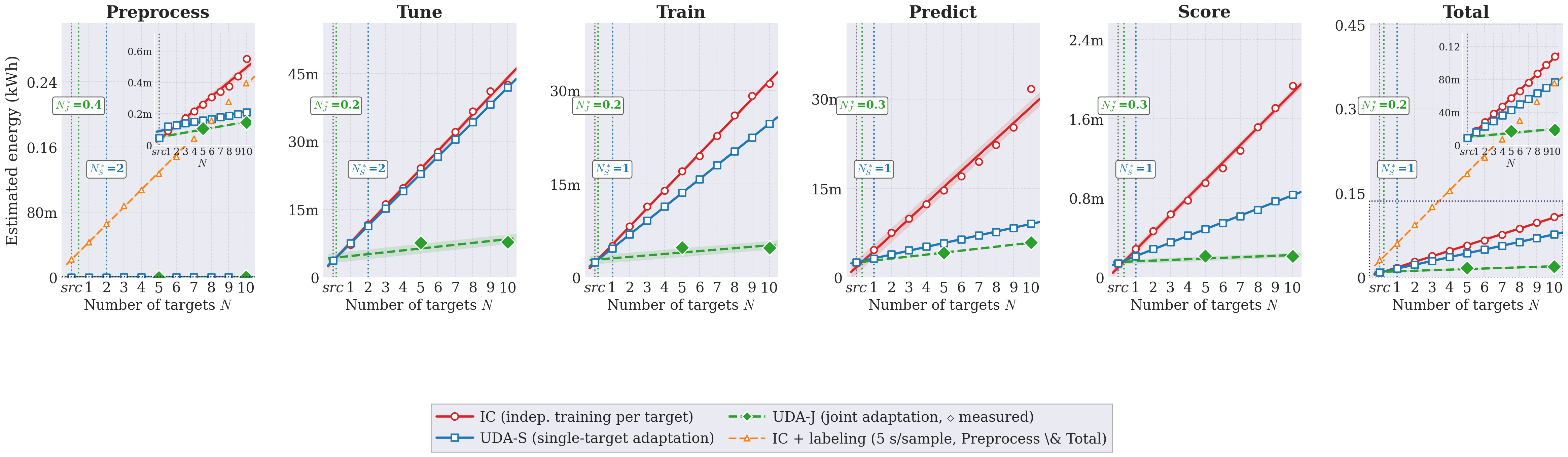}
  \caption{
    Estimated pipeline energy as a function of the number of target
    domains $N$ on the \textsc{HAR} dataset, broken down by pipeline stage and  aggregated (Total). \textbf{IC} trains one model per target independently; \textbf{UDA-S} adapts a shared source model sequentially; \textbf{UDA-J} performs a single joint adaptation ($\diamond$: measured at $N\!=\!5,10$).  }
  \label{fig:breakeven_har}
\end{figure*}
Figure~(b) shows a relative downward shift in accuracy compared
to~\citet{fawaz2025deep}, benchmark we adopt in this study. We attribute this gap to our reduced computational budget: whereas
~\citet{fawaz2025deep} uses up to thousands of tuning runs per model, our protocol is constrained to $5$ experiment sets within $15$--$400$ tuning and training budget range. This reflects a realistic deployment constraint and is itself a contribution of the study, but inevitably penalises tuning-sensitive models such as CoTMix and InceptionMix, whose wide interquartile ranges confirm high variance across runs. Overall, the energy-accuracy trade-off under our budget is best represented by Raincoat (low cost, robust accuracy) and InceptionRain (moderate cost, best accuracy among Inception-based UDA methods). The rest of our analysis focuses on these two UDA algorithms and InceptionTime for the IC setting.

\paragraph{Identifying the break-even phenomenon.}
Figure~\ref{fig:breakeven_har} addresses the central question of this work: \textit{when does UDA become the greener choice?} Each panel plots the estimated $E^{(c)}(N)$ as a function of the number of target domains $N$ for the three strategies: IC (red), UDA-S (blue), and UDA-J (green). The orange lines denoted as IC$+$labeling on the preprocessing and total panels correspond to $E_{\mathrm{label}}^{(c_1)}$, which accounts for the labeling cost.
Across all stages, the break-even values are remarkably low.
UDA-J reaches its threshold $N^*_{\mathrm{J}}$ at $N=1.0$-$1.4$
targets depending on the stage, meaning that adapting to a
joint target domain is already sufficient to amortise the source training cost $E_{\mathrm{source}}^{(c)}$.
UDA-S breaks even slightly later at $N^*_{\mathrm{S}} = 1$--$2$. 
Beyond these thresholds, both UDA strategies scale far more
favourably than IC: UDA-S grows linearly but with a slope
$E_{\mathrm{adapt}}^{(c)} \ll E_{\mathrm{ind}}^{(c)}$, while
UDA-J remains nearly flat as $E_{\mathrm{adapt,joint}}^{(c)}$
is fixed regardless of $N$. The Tune stage confirms that hyperparameter search is the dominant cost driver across all strategies. Both IC and UDA-S employ the same Hydra-based tuning protocol with identical budget settings, so their per-target tuning costs are comparable and both scale linearly with $N$.

The preprocessing and total panels show that once annotation cost
$E_{\mathrm{label}}^{(c_1)}$ is included, IC$+$labeling (orange)
diverges steeply from the computational-only IC curve (red),
making UDA strictly dominant from $N \geq 1$ in practical
scenarios where target labels are unavailable or expensive to
obtain. Under our experimental protocol, UDA becomes the greener choice from as few as $N^*=1$-$2$ target domains across every pipeline stage, and the advantage compounds rapidly with $N$.
When labeling cost is factored in, UDA is strictly preferable
from the very first target domain.

\section{Conclusion}

Task-specific factors in telecom settings, including labeling effort, dataset size, and cross-site heterogeneity, shape  scenarios in which UDA offers a clear technical advantage. However, the energy efficiency of UDA is not well studied, and it is important to evaluate it to ensure green communication networks. 
In this work, we present an energy--accuracy analysis of Unsupervised Domain Adaptation for time series classification, asking when UDA becomes the greener choice? Our break-even analysis shows that UDA reaches energy parity from as few as one to two target domains, and is strictly preferable from the first target once labeling cost is accounted for. Joint multi-target adaptation should be preferred whenever feasible.
Future work should extend evaluations to 
include more diverse datasets, 
develop unsupervised validation metrics to reduce tuning cost, and refine cost models under realistic labeling assumptions.

\section*{Acknowledgments}

\bibliography{biblio}
\bibliographystyle{plainnat}

\end{document}